\documentclass[conference]{IEEEtran}
\IEEEoverridecommandlockouts
\usepackage{cite}
\usepackage{amsmath,amssymb,amsfonts}
\usepackage{algorithmic}
\usepackage{graphicx}
\usepackage{textcomp}
\usepackage{xcolor}
\PassOptionsToPackage{hyphens}{url}\usepackage{hyperref}
\usepackage{tabularx}
\usepackage{multirow}
\usepackage{soul}
\usepackage{titlesec}
\usepackage{flushend}

\definecolor{myred}{HTML}{D83F31}
\definecolor{mygreen}{HTML}{186F65}

\makeatletter
    \newcommand{\linebreakand}{%
      \end{@IEEEauthorhalign}
      \hfill\mbox{}\par
      \mbox{}\hfill\begin{@IEEEauthorhalign}
    }
\makeatother

\def \triplebackticks{\textasciigrave\textasciigrave\textasciigrave}
\def \dropblock{\texttt{drop\_block()}}

\renewcommand{\baselinestretch}{0.99}

\begin{document}

\title{ChatGPT4PCG 2 Competition: Prompt Engineering for Science Birds Level Generation}

\author{
    \IEEEauthorblockN{Pittawat Taveekitworachai\IEEEauthorrefmark{1}, Febri Abdullah\IEEEauthorrefmark{2}, Mury F. Dewantoro\IEEEauthorrefmark{3}, Yi Xia\IEEEauthorrefmark{4},}
    \linebreakand
    \IEEEauthorblockN{Pratch Suntichaikul\IEEEauthorrefmark{5}, Ruck Thawonmas\IEEEauthorrefmark{6}, Julian Togelius\IEEEauthorrefmark{7}, and Jochen Renz\IEEEauthorrefmark{8}}
    \vspace{0.01cm}
    \linebreakand
    \IEEEauthorblockA{
        \textit{\{\IEEEauthorrefmark{1}\IEEEauthorrefmark{2}\IEEEauthorrefmark{3}\IEEEauthorrefmark{4}\IEEEauthorrefmark{5}Graduate School, \IEEEauthorrefmark{6}College\} of Information Science and Engineering}, \textit{Ritsumeikan University}, Kusatsu, Japan\\
    }
    \linebreakand
    \IEEEauthorblockA{
        \IEEEauthorrefmark{7}\textit{NYU Tandon School of Engineering}, \textit{New York University}, New York, USA
    }
    \linebreakand
    \IEEEauthorblockA{
        \IEEEauthorrefmark{8}\textit{School of Computing}, \textit{The Australian National University}, Canberra, Australia
    }
    \vspace{0.01cm}
    \linebreakand
    \IEEEauthorblockA{
    \{\IEEEauthorrefmark{1}gr0609fv, \IEEEauthorrefmark{2}gr0397fs, \IEEEauthorrefmark{3}gr0450xi, \IEEEauthorrefmark{4}gr0666ih, \IEEEauthorrefmark{5}gr0665kx\}@ed.ritsumei.ac.jp,\\
    \IEEEauthorrefmark{6}ruck@is.ritsumei.ac.jp, \IEEEauthorrefmark{7}julian@togelius.com, \IEEEauthorrefmark{8}jochen.renz@anu.edu.au
    }
}

\maketitle

\begin{abstract}
This paper presents the second ChatGPT4PCG competition at the 2024 IEEE Conference on Games. In this edition of the competition, we follow the first edition, but make several improvements and changes. We introduce a new evaluation metric along with allowing a more flexible format for participants' submissions and making several improvements to the evaluation pipeline. Continuing from the first edition, we aim to foster and explore the realm of prompt engineering (PE) for procedural content generation (PCG). While the first competition saw success, it was hindered by various limitations; we aim to mitigate these limitations in this edition. We introduce diversity as a new metric to discourage submissions aimed at producing repetitive structures. Furthermore, we allow submission of a Python program instead of a prompt text file for greater flexibility in implementing advanced PE approaches, which may require control flow, including conditions and iterations. We also make several improvements to the evaluation pipeline with a better classifier for similarity evaluation and better-performing function signatures. We thoroughly evaluate the effectiveness of the new metric and the improved classifier. Additionally, we perform an ablation study to select a function signature to instruct ChatGPT for level generation. Finally, we provide implementation examples of various PE techniques in Python and evaluate their preliminary performance. We hope this competition serves as a resource and platform for learning about PE and PCG in general\footnote{Source code and raw data: \url{https://bit.ly/cog-chatgpt4pcg2}}.
\end{abstract}


\section{Introduction}\label{sec:intro}
The first ChatGPT4PCG \cite{taveekitworachai2023chatgpt4pcg} saw great success in exploring the possibility of utilizing ChatGPT through prompt engineering (PE) \cite{white2023prompt} for Science Birds \cite{ferreira2014a} level generation. One PE technique named \textit{data store and retrieval} prompting was discovered through the first competition, in which ChatGPT is instructed to act as a data store. However, the first competition had some limitations. In particular, (1) the first competition's metrics were easily exploited through the generation of the same structures for each target character, (2) single-turn conversation prohibited the implementation of more advanced PE techniques recently introduced, and (3) a suboptimal image classifier and the choice of function signatures hindered ChatGPT from reaching its full potential.

Therefore, we introduce a new metric, \textit{diversity}, in this new edition of the competition to counteract a prompt designed to repeatedly generate the same structures for a particular target character, which hinder the usability of submitted solutions. Even though for a submitted prompt utilizing this repetitive structures approach to perform well, the structure for each target character must be carefully crafted through a certain technique, e.g., human experts, evolutionary algorithms, or other existing ML systems, we acknowledge that this does not align with the goals of the competition. Diversity is intended to emphasize that a generated structure not only needs to be stable and similar to the target character, but also diverse across all trials under the same target character. Diversity is calculated by computing an average of cosine distance between a pair of vector representations of each generated structure across trials for the same target characters of each prompt.

To support greater flexibility in implementing PE approaches for this competition, we change a submission format of the second competition. Instead of a single prompt text file in previous edition, we now accept a Python program. Accepting a program allows participants to implement various existing PE techniques that require multi-turn conversation, control flow, or external tools. Recently, various PE approaches require these capabilities for implementation. 

For example, least-to-most prompting \cite{zhou2023leasttomost} requires multi-turn conversation, tree-of-thought (ToT) prompting \cite{yao2023tree} requires the implementation of a tree traversal algorithm, such as breadth-first search (BFS) or depth-first search (DFS), and self-consistency \cite{wang2023selfconsistency} requires sampling multiple responses from LLMs. We also provide examples of an implementation for various PE approaches to help participants get started and as educational resources for those interested in PE. To ensure fairness, we impose some limitations on a number of tokens used and processing time for each trial using a newly developed Python package.

Another limiting factor of the previous competition was the image classifier used to assess the similarity of a generated structure to a target character. The classifier was trained on a publicly available dataset of handwritten characters. However, this was limited in the sense that generated Science Birds structures often do not resemble cursive writing as in handwritten characters. Therefore, we train a new image classifier based on a new dataset, generated from various selected fonts that resemble styles of Science Birds structures. Furthermore, we also conduct a study to select a better function signature for this edition of the competition, as this is another factor that can influence the performance of the model.

We hope that participants can explore and learn more about PE, which is one of the crucial skills for interacting effectively with LLMs and LLMs-integrated applications. Like the first competition, we also hope to gain a better understanding of PE for PCG and push the frontier of PE. Overall, our contributions are as follows:

\begin{itemize}
    \item We introduce ChatGPT4PCG 2, the second edition of ChatGPT4PCG competition, incorporating various improvements with the goal of educating and exploring PE for PCG.
    \item We introduce a platform for developing and evaluating PE approaches for Science Birds level generation. Additionally, we showcase examples implemented on this platform for various existing PE techniques.
    \item We perform and discuss experiments to confirm effectiveness of the changes we introduce for this edition of the competition.
\end{itemize}

\section{Related Work}\label{sec:related}
\subsection{Diversity for PCG Evaluation}\label{sec:related:chatgpt4pcg}
Diversity is an important metric in evaluating PCG systems. Several studies have discussed the need for reliable diversity metrics in different domains such as natural language generation \cite{tevet2021evaluating} and image synthesis \cite{han2022rarity}. These studies have highlighted limitations of existing metrics and have proposed new metrics that can effectively measure diversity. In particular, one study \cite{tevet2021evaluating} proposed a framework that measured the correlation between a diversity metric and a diversity parameter: the parameter used to control diversity in generated text. Similarly, another study \cite{han2022rarity} proposed the rarity score to measure the individual rarity of each image synthesized by generative models. These metrics aim to quantify the diversity or rarity degree of each generated output, allowing for better evaluation and comparison of PCG systems.

Another way to measure diversity is through the use of metrics such as entropy, which quantifies the amount of variety in the generated content \cite{abdullah2020generating}. A related approach is to use distance metrics to measure the dissimilarity between generated content items. High diversity is often desirable as it helps to avoid repetitive and predictable experiences for players, enhancing their enjoyment and engagement with the game \cite{gravina2019procedural}. Following the 2023 competition results, it is clear that embracing diversity as a key metric will not only improve the overall participant experience, but will also fuel a more innovative approach to PE for Science Birds level generation, encouraging participants to incorporate innovative ideas into their submitted approaches.

\subsection{PE}\label{sec:related:pe}
PE for LLMs is a rapidly evolving field, given its relatively novel emergence. Various PE techniques have been proposed recently and often go beyond single-turn conversation, requiring elaborate algorithm design. Basic PE techniques that could be applied within single-turn conversation, as in the previous edition of the competition, include zero-shot \cite{wei2022finetuned}, few-shot \cite{brown2020language}, and CoT prompting \cite{wei2022chain}, i.a. Nevertheless, more recent PE techniques utilizing multi-turn conversation and control flow have emerged, especially studies extending ideas of CoT prompting, including ToT, graph-of-thought (GoT) \cite{besta2023graph}, and everything-of-thought (XoT) prompting \cite{ding2023thoughts}. ToT prompting breaks a problem down into pre-defined smaller reasoning steps. For each step, multiple responses, i.e., thoughts, are sampled from an LLM and evaluated to choose the best one. Then follows the BFS or DFS to form a chain of thoughts. The ToT process then returns a final response.

GoT prompting extends ToT prompting by introducing concepts of refinement (loop) and aggregation. In GoT prompting, it is possible for an LLM to refine a thought by looping on the same thought, while aggregation merges multiple thoughts into one. With these two additional operations, the GoT process forms a directed graph; hence the name GoT. Finally, XoT prompting introduces an additional policy/value network for Monte-Carlo tree search (MCTS) to estimate a search structure, i.e., chains of thoughts, for an LLM to confirm and refine. This offers higher efficiency as it is cheaper to sample from a trained network through MCTS than having LLMs fully inference through the entire thought structure.

Therefore, we introduce support of program as a submission format in this competition to enable advanced PE techniques like ToT, GoT, and XoT prompting. We acknowledge that there are extensive studies that we did not mention here and worth exploring. We encourage participants to experiment and try various existing approaches or come up with their own novel PE techniques.

\section{ChatGPT4PCG 2}\label{sec:chatgpt4pcg2}
We update various rules because of the new changes introduced for ChatGPT4PCG 2. We change the function signature from \texttt{ab\_drop()} to \dropblock\ following our experiment as described in Section \ref{sec:experiment:function}, which is reflected in the rules detailed in Section \ref{sec:chatgpt4pcg2:evaluation:rules}. Additionally, we add/remove/update stages in the evaluation pipeline, detailed in Section \ref{sec:chatgpt4pcg2:evaluation}, where we also introduce diversity as a new metric. We also decide to replace \texttt{gpt-3.5-turbo-0613}, used by the previous competition, with \texttt{gpt-3.5-turbo-0125}, the latest available GPT-3.5 Turbo model from OpenAI as of writing, for the evaluation of this competition.

\subsection{Competition Evaluation}\label{sec:chatgpt4pcg2:evaluation}
We follow the majority of the evaluation stages from the previous competition. However, we introduce several changes to the evaluation pipeline compared to the previous competition to ensure compatibility with the updates introduced in this competition. A summary of the changes to the previous evaluation pipeline is as follows:

\begin{itemize}
    \item We remove the qualification checking script as this responsibility now falls under the purview of the new Python package and manual inspection of the submitted programs.
    \item The response gathering stage is changed from using the script introduced by the previous ChatGPT4PCG to manually running the submitted programs.
    \item The similarity checking script is modified to use the new classifier.
    \item We introduce an additional stage after similarity checking, which involves diversity checking using our newly developed diversity checking script. Further details about the script and the metric are discussed in Sections \ref{sec:chatgpt4pcg2:evaluation:pipeline} and \ref{sec:chatgpt4pcg2:evaluation:metrics}, respectively.
    \item The scoring-and-ranking script is modified to incorporate the new metric and adjustments described in Section \ref{sec:chatgpt4pcg2:evaluation:metrics}.
\end{itemize}

\subsubsection{\textbf{Rules of a Submitted Program}}\label{sec:chatgpt4pcg2:evaluation:rules}
While most of the rules remain similar to those of the previous competition, we update the function signature in the rules and removed outdated rules regarding the constraints of a submitted prompt text file from the previous competition. Additionally, we introduce new rules regarding the submitted program. We encourage our participants to review the exhaustive list of rules on our competition website\footnote{\url{https://chatgpt4pcg.github.io/rules}}. Below is a summarized version of the rules for the submitted program newly introduced in this competition:

\begin{itemize}
    \item A submitted program must interact with ChatGPT only via the provided APIs through our Python package to ensure a compatible output directory structure and fairness of the competition. Our Python package will monitor the amount of time and tokens used for each trial. Participants are responsible for ensuring that they use the latest version of our package upon submission.
    \item Any additional tools or services required as a part of an approach designed by participants must be provided to the organizers and ensure availability throughout the evaluation duration. Any paid tools or services are the responsibility of the participants to ensure availability to the organizers while complying with the terms, licenses, or agreements of each tool or service. We recommend participants consult additional information on the environment used for the evaluation to ensure compatibility of their program\footnote{\url{https://chatgpt4pcg.github.io/evaluation\#evaluation-env}}.
    \item The submitted program must not directly modify responses from ChatGPT to be written as a final output. Modification of the conversation history of ChatGPT with the intention of cheating is prohibited. Furthermore, modification of the Python package such as its token counter or timer used for the evaluation is also prohibited.
    \item An error during a trial is treated the same as producing an empty response.
    \item Each program's total size (including additional data and software to be downloaded as a part of the program instruction and environment) must not exceed 1GB. Each trial lasts only 120 seconds, and the maximum number of tokens used per trial is up to 25,000 tokens. The sampling temperature and random seed are always fixed at 1 and 42, respectively.
    \item We discourage the use of automatic prompt optimization during the evaluation process due to the token limit. However, we suggest utilizing these techniques for discovering optimal prompts to be used for the submitted program.
    \item Any programs that fail to follow the requirements are automatically disqualified.
\end{itemize}

\subsubsection{\textbf{Evaluation Pipeline}}\label{sec:chatgpt4pcg2:evaluation:pipeline}
The evaluation pipeline for ChatGPT4PCG 2 comprises eight stages. There are changes in all existing stages, with the addition of a diversity checking stage in this competition. Each submitted program goes through all of the stages. The stages are as follows:

\begin{enumerate}
    \item \textit{Qualification checking}: We manually inspect each submitted program for potential violations of the rules.
    \item \textit{Response gathering}: We manually run each submitted program to generate responses from ChatGPT for all target characters for a certain number of trials.
    \item \textit{Code extraction}: Similar to the previous competition, we use the code extraction script to extract a series of \dropblock\ function calls between the last pair of triple backticks (\texttt{\triplebackticks}).
    \item \textit{Text-to-XML conversion}: We convert the extracted code into an actual XML file describing a Science Birds level using the text-to-xml conversion script.
    \item \textit{Stability checking}: We evaluate the stability of each generated level using the Science Birds Evaluator introduced in the previous competition. The Evaluator also generates a picture of the generated structure using an all-black texture on a plain white background, following the same procedure as last year's.
    \item \textit{Similarity checking}: We utilize the similarity checking script introduced in the first competition, which incorporates an adjustment to use the new image classifier introduced in this competition. The rest of the functionalities remain the same as last year's.
    \item \textit{Diversity checking}: We introduce a new stage as a part of this year's evaluation pipeline. This stage utilizes a new script, the diversity checking script, assessing the diversity of generated structures for the same target characters using the same prompt across trials. The script averages the cosine distance of unordered all-pairs of outputs from a softmax function. More details about the calculation can be found in Section \ref{sec:chatgpt4pcg2:evaluation:metrics}.
    \item \textit{Scoring and ranking}: Same as last year, we utilize the scoring-and-ranking script for calculating the metrics. We adjust the script to incorporate changes we made to include the diversity metric.
\end{enumerate}

\subsubsection{\textbf{Evaluation Metrics}}\label{sec:chatgpt4pcg2:evaluation:metrics}
For the rest of this paper, $i$, $j$, and $k$ denote indices representing a particular trial, character, and program, respectively. $T$, $C$, and $P$ denote the total number of trials, characters, and programs, respectively. In this section, we describe three metrics used for the evaluation pipeline: stability ($sta_{ijk}$), similarity ($sim_{ijk}$), and diversity ($div_{jk}$) scores.

The calculations for $sta_{ijk}$ and $sim_{ijk}$ remain the same as the previous competition with only naming changes for better readability. On the other hand, the $div_{jk}$ is introduced in this competition to counter repetitive structures for the same target character by the same program. This metric encourages more creative and innovative approaches submitted by participants, with higher usability for scenarios outside of the competition. In other words, the metric discourages hard-coding structures. With the introduction of the new metric, we also adjust an equation used for calculating the weight of each character to include a component of the diversity score. Additionally, we multiply the diversity score by the character score of each team. The rest of this section discusses in more detail all metrics.

\begin{enumerate}
    \item \textbf{Stability}: The stability assessment approach of a generated level remains unchanged from the previous competition. The Evaluator still assesses a ratio of non-moving blocks against total blocks. However, we rename the stability score using the first three characters for better readability. Therefore, the stability score is defined as follows:
    $$
    sta_{ijk} = \frac{total\_blocks_{ijk} - moving\_blocks_{ijk}}{total\_blocks_{ijk}}
    $$
    \item \textbf{Similarity}: While the concept of the similarity score remains unchanged from the previous competition, we opt for a stricter definition of the softmax function, $\sigma$, where its output is a vector in order to support the introduction of $div_{jk}$. In combination with a new naming convention, we revise the equation as follows:
    $$
    sim_{ijk} = \sigma(\textbf{z}_{ijk})_j
    $$
    \item \textbf{Diversity}: In contrast to $sta_{ijk}$ and $sim_{ijk}$ that are calculated for each $ijk$, the diversity score is calculated across trials for $jk$ instead, as the diversity assessment requires multiple instances of generated levels. This score represents how diverse the generated levels are for the same target character using the same program. The $div_{jk}$ is calculated by computing the cosine distance, $D_c$, of unordered pairs in the set $A$ containing pairs of output vectors from the softmax function, $\textbf{v}_{ijk} = \sigma(\textbf{z}_{ijk})$. $\Xi_{jk}$ denotes a set containing all such vectors across trials of the same target character $j$ from the same program $k$. Therefore, $div_{jk}$ is defined as follows:
    $$
    div_{jk} = \frac{\sum_a^{|A_{jk}|} D_c(\textbf{v}_a^1, \textbf{v}_a^2)}{0.5T(T+1)-T}\text{,}
    $$
    where
    $$
    A_{jk} = \{(\textbf{v}_a^1, \textbf{v}_a^2) | (\textbf{v}_a^1, \textbf{v}_a^2)  \in \Xi_{jk} \bowtie \Xi_{jk} \land \textbf{v}_a^1 \neq \textbf{v}_a^2 \}
    $$
    Note that the denominator of $div_{jk}$ represents the maximum possible number of unordered pairs except for duplicates; a duplicate is $(\textbf{v}_a^1, \textbf{v}_a^2)$ where $\textbf{v}_a^1 = \textbf{v}_a^2$. This is to ensure that the range of possible values for $div_{jk}$ remains consistent with other scores, i.e., $[0, 1]$.
\end{enumerate}

Subsequently, the character weight ($weight_j$), trial score ($trial_{ijk}$), and character score ($char_{jk}$) are adjusted to accommodate the new naming convention and the addition of $div_{jk}$. For brevity, we only list the affected equations. For other equations and further details, readers may refer to the previous competition \cite{taveekitworachai2023chatgpt4pcg} or the competition's website\footnote{\url{https://chatgpt4pcg.github.io/evaluation}}.

\begin{itemize}
    \item \textbf{Character weight}:
    $$
    weight_{j} = w\_sta_{j} \times w\_sim_{j} \times w\_div_{j}\text{,}
    $$
    where
    $$
    w\_sta_{j} = max(1 - \frac{\sum_{k=1}^{P} \sum_{i=1}^{T} sta_{ijk}}{PT}, \frac{1}{C})\text{,}
    $$
    $$
    w\_sim_{j} = max(1 - \frac{\sum_{k=1}^{P} \sum_{i=1}^{T} sim_{ijk}}{PT}, \frac{1}{C})\text{,}
    $$
    and
    $$
    w\_div_{j} = max(1 - \frac{\sum_{k=1}^{P} div_{jk}}{P}, \frac{1}{C})
    $$
    \item \textbf{Trial score}:
    $$
    trial_{ijk} = weight_{j} \times sta_{ijk} \times sim_{ijk}
    $$
    \item \textbf{Character score}: We multiply $div_{jk}$ when calculating the character score because $div_{jk}$ is calculated across trials, as previously discussed. Similarly, $w\_div_j$ is averaged across programs only, not across trials and programs.
    $$
    char_{jk} = div_{jk}\frac{\sum_{i=1}^{T} trial_{ijk}}{T}
    $$
\end{itemize}

\section{Experiments}\label{sec:experiments}
This section introduces experiments performed to assess effectiveness of changes introduced in this edition of the competition. We study effectiveness of the improved image classifier in Section \ref{sec:experiment:vit}; the diversity as an additional metric in Section \ref{sec:experiment:diversity}; the new function signature in Section \ref{sec:experiment:function}; and various implementation of PE approaches in Section \ref{sec:experiment:pe}. In each experiment, we describe the setup of the experiment followed by their results and associated discussions.

\subsection{Effectiveness of the Improved Image Classifier}\label{sec:experiment:vit}
The image classifier plays an important role in the competition as the classifier is used for the similarity assessment. Furthermore, we also use results generated from the classifier for diversity assessment in this edition. Therefore, reliability of the classifier is crucial. We find that the old classifier poses one inherent limitation which is the dataset used to fine-tune the model. In the previous competition, they used a publicly available English handwritten letter dataset\footnote{\url{https://www.nist.gov/srd/nist-special-database-19}}. However, characteristics of handwritten characters are very different from a generated structure in Science Birds designed to resemble the character. The generated structure has a more uniform structure and shape compared to the handwritten variants. Therefore, we fine-tune a new classifier based on the same pre-trained model with a better dataset.

To the best of our effort, we could not find a publicly available dataset suitable for our purpose. Therefore, we create a new dataset by generating images of all 26 uppercase English characters from 99 fonts which we curate only fonts resembling the characteristics of the generated structures and have permissible licenses for our objective. In total, we have 2,574 generated images in our dataset. For each character, we divide it into 90 images for the training set and nine images for the testing set used during fine-tuning. We then follow the same procedure described in the previous competition for fine-tuning an open-source Vision Transformer (ViT) pre-trained model.

Nevertheless, the testing set from our new dataset is potentially subpar as a proxy for evaluating performance of generated structures in Science Birds given the constraints of the competition. Therefore, we prepare another evaluation set (\texttt{eval}) that reflects this objective. This evaluation set consists of manually constructed levels for all 26 uppercase English characters in Science Birds, where we also utilize the Evaluator used in the competition to generate images.

The manually constructed levels are obtained from the levels provided in a prompt submitted by `dereventsolve', the official winner of the previous competition and manually crafted by four co-authors. Each person is instructed to craft a Science Birds structure that closely resembles the target character as much as possible while maintaining the highest possible stability needed to construct such a level. In total, we have five evaluation images per character, totaling in 130 images for the evaluation set. We use the evaluation set to evaluate each model. The accuracy of the old and new ViT classifiers evaluated using the evaluation set are 0.4538 and \textbf{0.8231}, respectively.

We observe the new ViT classifier performs much better, almost doubling the performance on the evaluation set compared to the old classifier. This demonstrates the superiority of the new classifier as an evaluator in our use cases. Therefore, we employ the new classifier for this competition. Nevertheless, we acknowledge the limitation of the evaluation set due to its size, which stems from the fact that obtaining high-quality human-crafted levels is expensive. This underscores the importance of PCG and the main objective of this competition in discovering the best approach to automatically generate levels similar to the target character.

\subsection{Effectiveness of Diversity Metric}\label{sec:experiment:diversity}
We conduct an experiment to assess the effectiveness of the new diversity metric. We use participants' results from the previous competition but rerun them in two approaches. First, we rerun the evaluation pipeline starting from the new diversity checking stage until the end. This approach is denoted as ``Old ViT Classifier.'' Second, we rerun the evaluation pipeline starting from the similarity checking stage until the end. We denote the second approach as ``New ViT Classifier.'' We note that the first approach also provides advantages for all submissions, as some may be optimized to exploit the old classifier. We provide the original, ``Old ViT Classifier,'' and ``New ViT Classifier'' results in \autoref{tab:diversity}.

\begin{table*}[tbp]
\centering
\caption{Results from the previous competition are processed through the new evaluation pipeline, which includes diversity checking using oth the old and new ViT classifiers. The 2023 competition results (Original) are provided for comparison. $norm\_prompt_{k}$ and $prompt_{k}$ denote normalized prompt and prompt scores, respectively. \textbf{Rank} represents a final rank after the evaluation of each prompt where a sign in a pair of parentheses shows a trend of change compared to its original result.}
\begin{tabular}{r|ccr|ccr|ccr}
    \hline
    \multirow{2}{*}{\textbf{Prompt}} & \multicolumn{3}{c|}{\textbf{Original}} & \multicolumn{3}{c|}{\textbf{Old ViT Classifier}} & \multicolumn{3}{c}{\textbf{New ViT Classifier}}\\
    \cline{2-10}
    & \textbf{$norm\_prompt_{k}$} & \textbf{$prompt_{k}$} & \textbf{Rank} & \textbf{$norm\_prompt_{k}$} & \textbf{$prompt_{k}$} & \textbf{Rank} & \textbf{$norm\_prompt_{k}$} & \textbf{$prompt_{k}$} & \textbf{Rank}\\
    \hline
    \textbf{The Organizer} & \textbf{47.8425} & \textbf{0.3187} & \textbf{1} & \textbf{70.056} & \textbf{0.0577} & \textbf{1} (=) & \textbf{54.6391} & \textbf{0.0352} & \textbf{1} (=)\\
    dereventsolve & 31.1547 & 0.2076 & 2 & 0.0808 & 0.0001 & 12 (\textcolor{myred}{$\downarrow$}) & 4.1589 & 0.0027 & 4 (\textcolor{myred}{$\downarrow$})\\
    Soda & 4.7588 & 0.0317 & 3 & 8.496 & 0.007 & 2 (\textcolor{mygreen}{$\uparrow$}) & 11.3848 & 0.0073 & 3 (=)\\
    AdrienTeam & 3.3513 & 0.0223 & 4 & 7.9497 & 0.0065 & 3 (\textcolor{mygreen}{$\uparrow$}) & 15.7905 & 0.0102 & 2 (\textcolor{mygreen}{$\uparrow$})\\
    Saltyfish1884 & 2.1244 & 0.0142 & 5 & 3.9943 & 0.0033 & 4 (\textcolor{mygreen}{$\uparrow$}) & 1.4392 & 0.0009 & 9 (\textcolor{myred}{$\downarrow$})\\
    zeilde & 2.1233 & 0.0141 & 6 & 2.445 & 0.002 & 5 (\textcolor{mygreen}{$\uparrow$}) & 2.8 & 0.0018 & 6 (=)\\
    Team Staciiaz & 1.955 & 0.013 & 7 & 2.4223 & 0.002 & 6 (\textcolor{mygreen}{$\uparrow$}) & 3.2831 & 0.0021 & 5 (\textcolor{mygreen}{$\uparrow$})\\
    Harry Single Group & 1.8616 & 0.0124 & 8 & 1.6997 & 0.0014 & 7 (\textcolor{mygreen}{$\uparrow$}) & 2.5088 & 0.0016 & 7 (\textcolor{mygreen}{$\uparrow$})\\
    hachi & 1.5704 & 0.0105 & 9 & 0.5515 & 0.0005 & 10 (\textcolor{myred}{$\downarrow$}) & 0.732 & 0.0005 & 11 (\textcolor{myred}{$\downarrow$})\\
    Back to the future & 1.3773 & 0.0092 & 10 & 1.0352 & 0.0009 & 8 (\textcolor{mygreen}{$\uparrow$}) & 1.5972 & 0.001 & 8 (\textcolor{mygreen}{$\uparrow$})\\
    v1 (Baseline) & 1.1891 & 0.0079 & 11 & 0.8053 & 0.0007 & 9 (\textcolor{mygreen}{$\uparrow$}) & 0.9407 & 0.0006 & 10 (\textcolor{mygreen}{$\uparrow$})\\
    JUSTIN & 0.5243 & 0.0035 & 12 & 0.4642 & 0.0004 & 11 (\textcolor{mygreen}{$\uparrow$}) & 0.7248 & 0.0005 & 12 (=)\\
    Hope & 0.1488 & 0.001 & 13 & 0.0001 & 0 & 13 (=) & 0.0007 & 0 & 13 (=)\\
    albatross & 0.0162 & 0.0001 & 14 & 0 & 0 & 14 (=) & 0 & 0 & 14 (=)\\
    Prompt\_Wranglers & 0.0023 & 0 & 15 & 0 & 0 & 14 (\textcolor{mygreen}{$\uparrow$}) & 0 & 0 & 14 (\textcolor{mygreen}{$\uparrow$})\\
    For500 & 0 & 0 & 16 & 0 & 0 & 14 (\textcolor{mygreen}{$\uparrow$}) & 0 & 0 & 14 (\textcolor{mygreen}{$\uparrow$})\\
    \hline
\end{tabular}
\label{tab:diversity}
\end{table*}

The prompts `The Organizer' and `dereventsolve' employ \textit{data store and retrieval} prompting that could lead to the generation of repetitive structures. These prompts store pre-defined commands to build a structure for each character and instruct ChatGPT to behave like a data store, returning an associated value when queried with a key. In both cases, the key represents a target character. Despite having only one pre-defined structure per key, ChatGPT do not consistently generate the same structure for the same target character due to being a stochastic model. Therefore, the diversity score for each character of these prompts is not always zero. However, we observe that the penalty for diversity is applied appropriately, as `dereventsolve' is dropped from the second rank to twelfth using the old classifier and fourth using the new classifier.

There are two potential reasons why `dereventsolve' does not receive a significant penalty with the new classifier. First, the majority of prompts are unable to generate a response in the correct format, resulting in an automatic failure of such trials and receiving zero scores across all metrics. This is by design, as the competition aims for prompts to properly instruct ChatGPT to generate a response in the correct format. Therefore, prompts that fail to generate a level for a particular trial will receive a zero score for that trial. Second, levels provided by `dereventsolve' are potentially human-crafted and more closely resemble the shape of the character perceived by humans. Therefore, `dereventsolve' receives a higher standing using the new ViT classifier, which favors levels with higher resemblance to the character as perceived by humans. However, `dereventsolve' still suffers from the diversity penalty and cannot hold the second place in both evaluations.

Similar to `dereventsolve,' `The Organizer' also suffers from the diversity penalty, as evidenced by a significantly lower magnitude of $prompt_k$ when using the new classifier compared to the original. Furthermore, `The Organizer' also gets penalized for being optimized for the old classifier, in addition to the diversity penalty. However, `The Organizer' suffers less from the diversity penalty compared to `dereventsolve.' This could be due to the wording used for prompting the model to retrieve information, where the approach employed by `dereventsolve' results in more accurate retrieval compared to `The Organizer.' We believe that this stems from the characteristics of `dereventsolve,' which incorporates a compressed version of data and instructs ChatGPT to reconstruct it in a correct form, leading to more correct retrieval. Nevertheless, `The Organizer' holds the first place across all evaluations, likely due to similar reasons discussed previously for `dereventsolve,' where multiple prompts failed to generate responses in a correct format and are penalized, lowering their standing, unlike `dereventsolve' and `The Organizer.'

Excluding `The Organizer,' we find that `AdrienTeam' and `Soda' are the best-performing prompts. Both prompts generate levels with high stability and similarity scores while also maintaining a high diversity score for each target character. They are able to maintain their standing in either second or third places across the evaluations. Furthermore, they utilize few-shot prompting; this indicates that ChatGPT may need a few examples to efficiently perform the task given its complexity and various constraints. We recommend that participants further investigate the incorporation of few-shot prompting as part of their approach and optimizing their prompts to ensure that ChatGPT generates responses in the correct format.

\subsection{Function Signatures}\label{sec:experiment:function}
We perform an experiment to assess the effects of function signatures used for the competition, as variations in a prompt can lead to entirely different outcomes \cite{sclar2024quantifying}. We evaluate for 10 trials across all 26 characters. We prepare a total of four function signature candidates, including the original \texttt{ab\_drop(x, y)}. Each function signature is created while referring to Google's Python style guide\footnote{\url{https://google.github.io/styleguide/pyguide.html}}, with the intention of coming up with a more meaningful function signature and better conveying the functionalities of the function. We ask four graduate students in computer-related degree programs, each with prior programming experience in Python, to come up with three function signature candidates each. Later, we conduct an anonymous vote through an online service among these students, excluding one facilitator, to select the top three candidates with the highest vote count for this experiment. The four candidates, including the original, are as follows:

\begin{itemize}
    \item \textbf{fs1} denotes \texttt{drop\_block(block\_type: str, block\_position: int)}. This function signature indicates the functionality of dropping a block of type \texttt{block\_type} at a specified position, \texttt{block\_position}. We include type annotations to guide ChatGPT in providing arguments of correct types.
    \item \textbf{fs2} denotes \texttt{drop\_block(block\_type: str, x\_position: int)}. \textbf{fs2} is similar to \textbf{fs1} with a slight change to the second parameter name, indicating a position of dropping a block along the x-axis.
    \item \textbf{fs3} denotes \texttt{falling\_block(type: str, x\_position: int)}. \textbf{fs3} is designed differently, describing the result of the function instead.
    \item \textbf{original} denotes the original \texttt{ab\_drop(x, y)}.
\end{itemize}

For the base prompt, we use a baseline prompt, \texttt{v1}, from the previous competition and modify all places mentioning the function names, parameters/arguments, and signatures, accordingly. We use the evaluation pipeline introduced in this study, including the diversity checking stage, for this experiment. \autoref{tab:function_signature} presents the experiment results.

\begin{table}[tbp]
\centering
\caption{Four function signature candidates, including the original from the previous study, are evaluated using the pipeline introduced in this study to determine the most suitable one. Among all four candidates, \textbf{fs2} demonstrates the best performance.}
\begin{tabular}{rccc}
    \hline
    \textbf{Function signature} & \textbf{$norm\_prompt_{k}$} & \textbf{$prompt_{k}$} & \textbf{Rank}\\
    \hline
    \textbf{fs2} & \textbf{69.7537} & \textbf{0.0002} & \textbf{1}\\
    fs1 & 30.2463 & 0.0001 & 2\\
    fs3 & 0 & 0 & 3\\
    original & 0 & 0 & 3\\
    \hline
\end{tabular}
\label{tab:function_signature}
\end{table}

We observe that \textbf{fs2} outperforms the other candidates. The performance discrepancy highlights the significant impact of function signature design on model performance, aligning with findings from previous studies \cite{sclar2024quantifying,lu2022fantastically}. These studies indicate that LLM performance is sensitive to prompt wording, formatting, and content organization in a prompt. Notably, both \textbf{fs3} and the \textbf{original} fail to generate correctly formatted responses, resulting in zero scores for both programs. This suggests that \textbf{fs1} and \textbf{fs2} provide ChatGPT with clearer context regarding the function's operation, compared to emphasizing the result of executing that function.

Comparing \textbf{fs1} and \textbf{fs2}, which have similar content, we observe that ChatGPT may favor a function signature with a more straightforward parameter name. Specifically, \texttt{x\_position} in \textbf{fs2} offers better contextual understanding compared to \texttt{block\_position} in \textbf{fs1}. The latter parameter name could be ambiguous regarding its meaning without delving into the function's inner workings regarding the axis for a dropping position. Therefore, \texttt{x\_position} in \textbf{fs2} provides clearer expectations for ChatGPT and explicitly conveys how this parameter influences the function.

\subsection{Effective of PE Examples}\label{sec:experiment:pe}
We provide examples of how to implement selected PE techniques. All techniques are evaluated in the same manner as described in Section \ref{sec:experiment:function}. We select seven PE techniques: five basic techniques and two techniques requiring control flow. The five basic PE techniques are (1) zero-shot, (2) null-shot, (3) few-shot, (4) zero-shot CoT, and (5) null-shot CoT prompting. The two PE techniques requiring control flow are (6) zero-shot with multi-turn conversation and (7) ToT prompting. For all techniques, the base prompts are \texttt{v1}, the baseline prompt of the previous competition. We make modifications and/or incorporate additional prompts as needed.

For zero-shot prompting, the prompt is the same as \textbf{fs2} described in Section \ref{sec:experiment:function}. Null-shot prompting adds an additional phrase, obtained from a study \cite{taveekitworachai2024large} exploiting hallucination in LLMs for enhanced performance, at the end of the zero-shot prompt. For few-shot prompting, we add one additional section called ``Examples'' at the end of the zero-shot prompt consisting of one example for each of the three hardest characters, i.e., characters with the highest weights, according to the results of the previous competition, which are `G,' `Q,' and `S,' ordered alphabetically. An example of each character is selected manually using convenience sampling from \texttt{eval}, described in Section \ref{sec:experiment:vit}, due to the lack of a broader dataset for this task. Zero-shot CoT prompting adds an additional phrase, obtained from \cite{kojima2022large}, at the end of the zero-shot prompt, enabling the model to rely on its own reasoning abilities and perform the task according to its own reasoning steps. Similarly, we add one additional phrase detailed in the original study of null-shot CoT prompting \cite{taveekitworachai2024large} at the end of the zero-shot prompt.

For zero-shot prompting with multi-turn conversation, the first turn is the same as zero-shot prompting, while the second turn is a format-guiding instruction to guide the model in outputting a correct format as required by the competition. Finally, ToT prompting is the most elaborate and complex technique we select for this experiment. We choose a BFS variant instead of a DFS variant, as we believe that a final output may benefit from greater exploration at each step rather than relying more on exploitation. This decision is made partly due to the complexity of the task and the existence of multiple valid solutions. Nevertheless, participants interested in DFS are encouraged to further explore. We also limit both the maximum depth, i.e., the number of reasoning steps, and the branching factor, i.e., the number of thought candidates, to two to simplify the experiment.

We adhere to the style of ToT prompting \cite{yao2023tree} to develop prompts necessary for this approach. For a task prompt, we use a modified version of the zero-shot prompt, incorporating additional information about reasoning steps for this task, along with an example of the task for character ``A,''. To evaluate each thought, we leverage ChatGPT by developing another evaluation prompt that outputs two numbers indicating the quality of a thought in terms of stability and similarity. A final output prompt, used for composing the intermediate thoughts along the chain, is the format-guiding sentence, similar to the one used in zero-shot prompting with multi-turn conversation.

We note that the techniques selected for this experiment are not exhaustive. Furthermore, the prompts and hyperparameters utilized for each PE technique in this experiment may not be optimal. We hope that this experiment serves as a starting point for participants to experiment with and further improve upon, and as a great resource for learning PE in general. For basic techniques, we select each technique to demonstrate how variations in the prompts could influence the outcomes, while the other two techniques are selected to illustrate how participants may implement a PE program for multi-turn conversation and complex algorithm design, such as in ToT prompting. The results of the experiment are shown in \autoref{tab:pe}.

\begin{table}[tbp]
\centering
\caption{Results when each PE technique competes with each other, demonstrating preliminary analyses of each PE for the level generation task.}
\begin{tabular}{rccc}
    \hline
    \textbf{PE} & \textbf{$norm\_prompt_{k}$} & \textbf{$prompt_{k}$} & \textbf{Rank}\\
    \hline
    \textbf{Zero-shot (multi-turn)} & \textbf{46.5437} & \textbf{0.0049} & \textbf{1}\\
    Few-shot & 26.3907 & 0.0028 & 2\\
    ToT & 11.412 & 0.0012 & 3\\
    Null-shot CoT & 8.0634 & 0.0008 & 4\\
    Null-shot & 5.3641 & 0.0006 & 5\\
    Zero-shot CoT & 2.0619 & 0.0002 & 6\\
    Zero-shot & 0.1643 & 0 & 7\\
    \hline
\end{tabular}
\label{tab:pe}
\end{table}

Zero-shot prompting with multi-turn conversation shows the best performance among the others. This result highlights an interesting observation: this technique ranks at the top, while the original zero-shot prompting positions at the last place. This suggests that improper formatting is the primary obstacle hindering the performance of the baseline prompt from the previous competition.

This best-performing approach incorporates a format-guiding sentence as a final step for evaluation, resulting in a higher rate of ChatGPT generating in the correct format. Therefore, these results confirm the effectiveness of the format-guiding sentence for this task and suggest that it should be incorporated into approaches to achieve a higher score. We recommend that participants incorporate a similar tactic into their approaches for better performance.

Results obtained using few-shot prompting, ranks second, reveal interesting observations. ChatGPT generally performs well only for the characters with examples provided. However, we also observe that this approach also performs well for the character ``I,'' where ChatGPT stacks square blocks, \texttt{b11}, vertically to construct the shape of ``I,'' even without seeing such an example. This suggests the possibility that with appropriate few-shot examples included in the prompt, ChatGPT may be able to understand the task through \textit{in-context learning} \cite{brown2020language}. Alternatively, another approach to further improve performance may involve providing better explanations in combination with the provided examples.

ToT prompting ranks third in performance. This demonstrates the potential of ToT prompting, which may achieve higher performance with improved prompts and hyperparameters. Additionally, this result indicates the potential of similar approaches in dividing a problem into reasoning steps, with exploration aiding ChatGPT's performance. We speculate that similar approaches, including GoT and XoT prompting, may also show promising results.

\section{Conclusions}\label{sec:conclusions}
We present ChatGPT4PCG 2 at IEEE CoG 2024, building upon the success of the first competition in the hope of sparking interest in PE and PCG. This competition introduces several changes to enhance the experience of participants and the fairness of the competition. One significant addition is the introduction of the diversity score as a new metric to discourage the generation of repetitive levels. In addition, we improve the image classifier used for the similarity assessment. Furthermore, we also change the submission format to a Python program, providing greater flexibility in implementing state-of-the-art PE techniques or devising new ones.

Numerous experiments are conducted to validate the effectiveness of the changes introduced in this edition. We also provide an initial assessment of various PE techniques. Moreover, we open-source all the implemented PE programs to promote education in PE and its applications in PCG. Similar to the first competition, we hope that our competition sparks interest not only in PE, which is becoming an increasingly important skill for effectively interacting with LLMs, but also in PCG, pushing the boundaries of both fields. Like the first competition, we plan to report results and findings obtained from this competition to facilitate further discussion and advancements in PE and PCG.

\bibliographystyle{IEEEtran}
\bibliography{refs}

\end{document}